\documentclass[preprint,12pt]{elsarticle}

\usepackage{amssymb}
\usepackage{amsmath} 
\DeclareMathOperator{\Tr}{Tr}
\usepackage{graphicx} 
\usepackage{stmaryrd}
\usepackage{listings}
\usepackage{xcolor}
\usepackage{algorithm}
\usepackage{algpseudocode}
\usepackage{placeins}
\usepackage{mathtools}

\journal{Computer Methods in Applied Mechanics and Engineering}

\begin{document}

\begin{frontmatter}

\title{Exploring energy minimization to model strain localization as a strong discontinuity using Physics Informed Neural Networks}

\author[inst1,inst3]{Omar León}
\author[inst3]{Víctor Rivera}
\author[inst1,inst3,inst4]{Angel Vázquez-Patiño}
\author[inst5,inst6]{Jacinto Ulloa}
\author[inst2,inst3]{Esteban Samaniego}

\affiliation[inst1]{organization={Department of Civil Engineering, University of Cuenca},
            city={Cuenca},
            postcode={010203}, 
            state={Azuay},
            country={Ecuador}}

\affiliation[inst2]{organization={Departamento de Recursos Hídricos y Ciencias Ambientales, University of Cuenca},
            city={Cuenca},
            postcode={010207}, 
            state={Azuay},
            country={Ecuador}}
\affiliation[inst3]{organization={Faculty of Engineering, University of Cuenca},
            city={Cuenca},
            postcode={010207}, 
            state={Azuay},
            country={Ecuador}}

\affiliation[inst4]{organization={Faculty of Architecture, University of Cuenca},
            city={Cuenca},
            postcode={010203}, 
            state={Azuay},
            country={Ecuador}}

\affiliation[inst5]{organization={Department of Mechanical Engineering, University of Michigan},
            city={Ann Arbor},
            postcode={48109}, 
            state={Michigan},
            country={United States of America}}      
            
\affiliation[inst6]{organization={Division of Engineering and Applied Science, California Institute of Technology},
            city={Pasadena},
            postcode={91125}, 
            state={California},
            country={United States of America}}            

\begin{abstract}
We explore the possibilities of using energy minimization for the numerical modeling of strain localization in solids as a sharp discontinuity in the displacement field. For this purpose, we consider (regularized) strong discontinuity kinematics in elastoplastic solids. The corresponding mathematical model is discretized using Artificial Neural Networks (ANNs), aiming to predict \emph{both} the magnitude and location of the displacement jump from energy minimization, \textit{i.e.},  within a variational setting. The architecture takes care of the kinematics, while the loss function takes care of the variational statement of the boundary value problem. The main idea behind this approach is to solve both the equilibrium problem and the location of the localization band by means of trainable parameters in the ANN. As a proof of concept, we show through both 1D and 2D numerical examples that the computational modeling of strain localization for elastoplastic solids using energy minimization is feasible. 
\end{abstract}


\end{frontmatter}


\section{Introduction}
\label{sec:sample1}
Strain localization is a material instability that entails the appearance of zones within a solid in which strains are highly concentrated. It is a material instability in the sense that it can occur even in regions in which the stress is uniform. Typically, strain localization is related to constitutive models with softening. It is a general phenomenon associated with different types of failure mechanisms like fracture in brittle and quasi-brittle materials, or shear bands in geomaterials and metals. Hence, its study is of enormous importance from the solid mechanics point of view.

The modeling of strain localization in solids is a very complex task. Standard boundary value problems that model the behavior of solids become ill-posed when strain localization is observed. This is manifested in the tendency of the model to dissipate energy in bands whose width tends to zero. These localized dissipative mechanisms are not easy to capture in standard formulations within the continuum mechanics framework. When the mathematical models are discretized, this ill-posedness is inherited by the numerical model in the form of spurious behavior. Typically, if no regularization is introduced, the dissipation tends to zero as the discretization is refined, which is not what is observed in reality. This lack of objectivity makes the results obtained by the numerical model useless in the post-critical stages. 

To solve this lack of objectivity two main strategies have been followed in the literature. On the one hand, for rate-independent models, an intrinsic characteristic length scale is introduced in the constitutive law so that the width of a localization band does not tend to zero. A very common option for this strategy is to use higher-order terms in the governing equations. The intrinsic length scale can also be introduced by adding a scalar field to describe the location of the localization band as a smooth transition between the parts of the material that are localized and those that are not. Then, the band has finite width and the kinematic fields (displacement and strain) do not have discontinuities. In this context, phase fields have been used extensively to simulate fracture with regularized kinematics \cite{bourdin2008variational, miehe2010phase, pham2011gradient, ulloa2019phase, rodriguez2018variational}.  

On the other hand, strain localization has been approached with strategies that incorporate strong discontinuity kinematics in the displacement field (\textit{e.g.}, \cite{article}). Examples of these methods are the strong discontinuity approach (SDA) \cite{simo1993analysis} and the eXtended Finite Element Method (XFEM) \cite{moes2002extended,samaniego2005continuum,vigueras2015xfem}. Such approaches have given good results, although they have shown problems when determining both the onset of the localization band and its location.

Regarding the equilibrium of the solid, one particularly powerful option is to use a variational framework. In fact, the most popular phase-field formulations currently used are devised in this framework. Following the work of Francfort and Marigo \cite{francfort1998revisiting}, in which fracture was modeled as a discontinuity in the displacement field, Bourdin \textit{et al.} \cite{bourdin2008variational} proposed a phase-field formulation as an approximation of the problem with a sharp discontinuity.  The variational setting has proved to be a rigorous one \cite{samaniego2021variational,bourdin2008variational,del2013diffuse}. In our contribution, we are closer to the original treatment of Francfort and Marigo~\cite{francfort1998revisiting}: we consider a variational framework for free discontinuity problems, that is, maintaining kinematics that allows for sharp discontinuities, without introducing a localization limiter in the formulation, that is, without a phase field. In addition, we rely on concepts from the so-called strong discontinuity approach \cite{OLIVER20007207} both for the kinematics and to ensure objectivity. 

Many options to discretize a boundary value problem (BVP) are available. Perhaps the most popular one in solid mechanics is the Finite Element Method (FEM). Recently, a promising alternative has emerged: the use of artificial neural networks (ANNs) to generate approximation spaces to solve a BVP, \textit{i.e.}, the so-called Physics-Informed Neural Networks (PINNs) \cite{raissi2019physics}. It has been recently shown how ReLU-PINNs can emulate the FEM approximation space \cite{he2018relu}. For instance, He \textit{et al.} \cite{he2018relu} demonstrated how an ANN using Continuous Piece-Wise Linear (CPWL) functions can generate the finite element basis functions.

The computational implementation of the variational/energy approach to solid mechanics is particularly interesting in the context of PINNs. A PINN is based on the implementation of a slightly modified ANN. Indeed, the training mechanics of a PINN are basically the same as that of any ANN: it is based on a network architecture and an optimizer, which uses automatic differentiation to calculate gradients \cite{griewank2008evaluating}. The fundamental novelty is to consider a loss function that contains information on physical phenomena \cite{raissi2019physics, Katsikis2022, bai_introduction_2023}. In general, the loss function is formulated based on the residuals of the governing equations. In this sense, balance equations are introduced as constraints (in analogy with the concept of data-driven computing~\cite{kirchdoerfer2016data}). However, for some mechanical problems, it seems natural to use a variational approach \cite{samaniego2021variational} involving energy minimization. This approach was presented by Weinan and Yu \cite{weinan2018deep} as the Deep Ritz Method (DRM). Within this framework, Samaniego \textit{et al.} \cite{samaniego2020energy} presented examples for many problems in solid mechanics including the phase-field modeling of fracture. This approach, called the deep energy method, has been used for several applications such as hyperelasticity \cite{nguyen2020deep} or gradient elasticity \cite{nguyen2021parametric}. Further contributions focusing on phase-field fracture models can be found, for instance, in \cite{goswami2020transfer} and, more recently, \cite{manav2024phase}.

Conversely, attempts to model strain localization as a sharp discontinuity using the variational statement of the BVP and discretized by means of Neural Networks (NNs) are not found in the literature. Perhaps the only exception is \cite{baek2022neural}; however, sensu stricto, this interesting proposal uses a Reproducing Kernel Particle Method enhanced with an NN.

Motivated by this knowledge gap, our main objective is to explore the possibility of modeling strain localization in elastoplastic solids within an energetic framework using NNs as the only approximation space for the displacement field, including (regularized) discontinuities. In particular, we aim to predict, naturally from the variational principle, both the magnitude of the displacement jump and the location of the localization band, addressing a persistent challenge in advanced finite element technologies such as SDA or XFEM. To this end, we resolve both the equilibrium problem and the location of the localization band using trainable parameters in an ANN. Specifically, the architecture takes care of the strong discontinuity kinematics while the loss function takes care of the variational statement of the BVP. In this way, it might be said that, in some sense, we are solving a free discontinuity problem numerically without resorting to regularizing gradients or a new field (\textit{e.g.}, a phase field). As a proof of concept, we show, via both 1D and 2D numerical examples, that the computational modeling of strain localization as a sharp discontinuity in elastoplastic solids using energy minimization is~feasible.

\section{A free discontinuity problem}

From a macroscopic point of view, the localization band can be described as a jump or discontinuity surface of the displacement field, which is called in the computational mechanics community a strong discontinuity \cite{Manzoli1999}.

Within the framework of the so-called free discontinuity problems \cite{ambrosio2013variational}, this kinematics is related to an energy functional $\mathcal{W}$ that can be additively decomposed into a bulk energy $\mathcal{E}$ and a surface energy $\mathcal{S}$ \cite{samaniego2021variational}:
\begin{equation}
\label{energy_free discontinuity}
\mathcal{W}[\boldsymbol{u},\Gamma]=\mathcal{E}[\boldsymbol{u}, \Gamma]+\mathcal{S}[\llbracket \boldsymbol{u} \rrbracket,\Gamma],
\end{equation}
where $\llbracket \boldsymbol{u} \rrbracket$ is the jump in displacement field $\boldsymbol{u}$ at the discontinuity surface $\Gamma$. In free discontinuity problems, both $\boldsymbol{u}$ and $\Gamma$ are unknowns.

Here we consider an approximation of this surface by a band of constant width $h$ centered at $\Gamma$: 
\begin{equation}
\Gamma_{h} \approx \Gamma. 
\end{equation}
This implies considering a regularization of the strong discontinuity kinematics. It is important to mention that $h$ is, in principle, arbitrarily small and that, when $h$ goes to zero, the jump discontinuity at $\Gamma$ is recovered.

\section{Kinematics of regularized strong discontinuities}

\begin{figure}[b]
\centering
\includegraphics[width=0.7\textwidth]{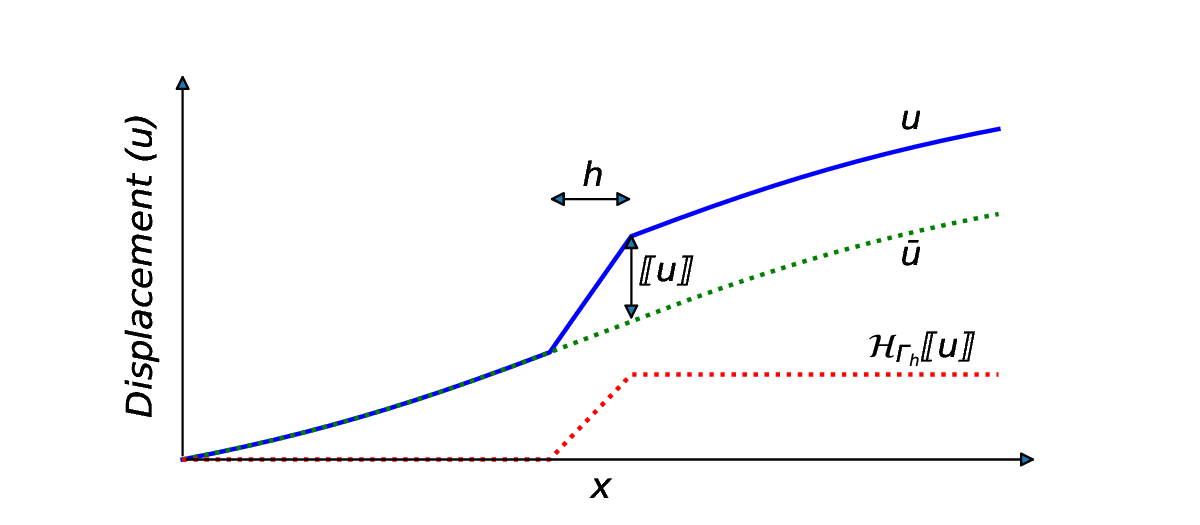}
\caption{Kinematic decomposition for a one-dimensional problem. A strong discontinuity is recovered as $h$ tends to zero.}
\centering
\label{fig_dec}
\end{figure}

A strong discontinuity can be regularized in the form of a so-called weak discontinuity, where the width of the localization band is no longer equal to zero, but a small positive parameter $h$. Consider the following additive decomposition of the displacement field, $\boldsymbol{u}$ (Figure~\ref{fig_dec}):
\begin{equation}\label{eq8}
\boldsymbol{u}=\Bar{\boldsymbol{u}} +  \mathcal{H}_{\Gamma_{h}} \llbracket \boldsymbol{u} \rrbracket,
\end{equation}
where $\Bar{\boldsymbol{u}}$ is the continuous part of the displacement, $\mathcal{H}_{\Gamma_{h}}$ is a regularized Heaviside that varies linearly across $\Gamma_{h}$, and $\llbracket \boldsymbol{u} \rrbracket$ is the jump~\cite{article,simo1993analysis}. From this decomposition and defining the strain as the symmetric gradient of the displacement, one obtains the  additive decomposition of the strain tensor
\begin{equation}
\boldsymbol{\epsilon} = \nabla^s \boldsymbol{u} =\hat{\boldsymbol{\epsilon}} + \frac{1}{h} \mu_{\Gamma_{h}} \llbracket \boldsymbol{u} \rrbracket \odot \boldsymbol{n},
\end{equation}
where $\hat{\boldsymbol{\epsilon}}$ is the regular part of the strain, $\mu_{\Gamma_{h}}$ is a collocation function acting in $\Gamma_{h}$, and $\boldsymbol{n}$ is the unit normal to the discontinuity surface $\Gamma$.

To model plasticity, we consider the additive decomposition of the strain:
\begin{equation}
\label{AddDecomp}
\boldsymbol{\epsilon}=\boldsymbol{\epsilon}^{e}+\boldsymbol{\epsilon}^{p}.
\end{equation}
We also define the equivalent plastic strain $p$ as usual:
\begin{equation}
p= \int_0^t \Vert \dot{\boldsymbol{\epsilon}}^{p}\Vert \, d\tau.
\end{equation}
Notice that
\begin{equation}
\dot{\boldsymbol{\epsilon}}^{p}=\dot{p} {\boldsymbol{m}},
\end{equation}
where $\boldsymbol{m}$ is the so-called plastic flow direction.

For simplicity, we will assume here that 
\begin{equation}
\label{EpEqEj}
\dot{\boldsymbol{\epsilon}}^{p}=\frac{1}{h} \mu_{\Gamma_{h}} \llbracket \dot{\boldsymbol{u}} \rrbracket \odot \boldsymbol{n}.
\end{equation}
This implies that plastic flow and localized strains co-evolve within $\Gamma_h$, while the regular strains in the bulk evolve elastically. In addition, if one assumes that $p=0$ at $t=0$ and that plasticity, once initiated, evolves monotonically without elastic unloading, then
\begin{equation}
\label{pEqJump}
p= \frac{1}{h} \mu_{\Gamma_{h}} \Vert {\llbracket \boldsymbol{u} \rrbracket} \Vert .
\end{equation}
As a consequence, 
\begin{equation}
\boldsymbol{\epsilon}^{p}=p {\boldsymbol{m}}=\frac{1}{h} \mu_{\Gamma_{h}} \llbracket \boldsymbol{u} \rrbracket \odot \boldsymbol{n}.
\end{equation}
We will also consider that plastic flow is isochoric:
\begin{equation}
\label{Isochoric}
\Tr(\boldsymbol{\epsilon}^{p})=\frac{1}{h} \mu_{\Gamma_{h}} \llbracket \boldsymbol{u} \rrbracket \cdot \boldsymbol{n} = 0 .
\end{equation}
Notice that this implies that the jump is orthogonal to the unit normal to $\Gamma$, as in mode II fracture.

\section{Variational modeling of elastoplastic solids}
\subsection{Variational formulation}
We consider the following functional for an elastoplastic solid occupying a volume $\Omega$:
\begin{equation}
\label{energyfunct_plasticity}
\begin{aligned}
\mathcal{W}_h[\boldsymbol{u},p]
=&\int_{\Omega}\Psi^e(\boldsymbol{\epsilon}^e(\boldsymbol{u},\boldsymbol{\epsilon}^p)) \, dV +\int_{\Omega}\bigg(\frac{1}{2}H p^2 + \sigma_{p} \, p \bigg) \, dV , \\
\end{aligned}
\end{equation}
where $\Psi^e$ is the elastic energy density, $H$ is the softening parameter, and $\sigma_{p}$ is the yield stress. Noting that the Cauchy stress follows from the usual Coleman-Noll procedure as
\begin{equation}
    \boldsymbol{\sigma}=\frac{\partial\Psi^e}{\partial\boldsymbol{\epsilon}^e},
\end{equation}
one may show~\cite{ulloa2021variational,alessi2015gradient} that an energy functional of this type corresponds to a plasticity model governed by the standard KKT triplet
\begin{equation}
    f(\boldsymbol{\sigma})\leq0,\quad \gamma\,f(\boldsymbol{\sigma})=0, \quad \gamma\geq0,
\end{equation}
and the associative flow rule
\begin{equation}
    \dot{\boldsymbol{\epsilon}}^p = \gamma \, \partial f (\boldsymbol{\sigma}), \quad \dot{p}= k_N\,\gamma,
\end{equation}
where the yield function reads
\begin{equation}
    f(\boldsymbol{\sigma}) = \Vert \mathrm{dev} \, \boldsymbol{\sigma}\Vert - k_N\,(\sigma_p + H\,p).
\end{equation}
Here,
\begin{equation}
    k_N = 
    \begin{dcases} 1  & \text{if $N=1$}, \\
    \frac{\sqrt{N-1}}{N} & \text{otherwise},
    \end{dcases}
\end{equation}
where $N\in\{1,2,3\}$ is the spatial dimension.
Note that non-associative plasticity models, often useful for geomechanics or cyclic loading, may also be considered within an extended variational framework~\cite{ulloa2021variational,ulloa2021phase}.

\subsection{Incorporating jumps}

From Equation \eqref{pEqJump}, the energy functional can be written as 
\begin{equation}
\label{energyfunct_plasticity2}
\begin{aligned}
\mathcal{W}_h[\boldsymbol{u},p]
=&\int_{\Omega}\Psi^e(\boldsymbol{\epsilon}^e(\boldsymbol{u},
\boldsymbol{\epsilon}^p)) dV + \int_{\Gamma_h} \bigg( \frac{1}{2h} \Bar{H} \Vert \llbracket \boldsymbol{u} \rrbracket \Vert^2 + \frac{1}{h} \sigma_{p} \Vert {\llbracket \boldsymbol{u} \rrbracket} \Vert \bigg) \, dV ,\\
\end{aligned}
\end{equation}
where $\Bar{H}$ is given by
\begin{equation}
H=h \Bar{H}.
\end{equation}
The parameter $\Bar{H}$ is termed intrinsic softening parameter and is considered a material property. Notice that, when the bandwidth $h$ tends to zero, we obtain
\begin{equation}
\label{energyfunct_cohesive}
\begin{aligned}
\mathcal{W}[\boldsymbol{u},p]
=&\int_{\Omega}\Psi^e(\boldsymbol{\epsilon}^e(\boldsymbol{u},
\boldsymbol{\epsilon}^p)) dV + \int_{\Gamma} \bigg( \frac{1}{2} \Bar{H} \Vert \llbracket \boldsymbol{u} \rrbracket \Vert^2 + \sigma_{p} \Vert {\llbracket \boldsymbol{u} \rrbracket} \Vert \bigg) \, dS .\\
\end{aligned}
\end{equation}
As a consequence, the last two terms of Equation \eqref{energyfunct_cohesive} tend to surface energies on $\Gamma$. From this, we can see that a cohesive energy density is induced:
\begin{equation}
\label{cohesive_energy}
\begin{aligned}
\psi(\llbracket \boldsymbol{u} \rrbracket)=\frac{1}{2} \Bar{H} \Vert \llbracket \boldsymbol{u} \rrbracket \Vert^2
 + \sigma_{p} \Vert {\llbracket \boldsymbol{u} \rrbracket} \Vert .\\
\end{aligned}
\end{equation}
The corresponding cohesive force (per unit surface) reads
\begin{equation}
\label{cohesive_force}
\begin{aligned}
\boldsymbol{t}_c(\llbracket \boldsymbol{u} \rrbracket)= \sigma_{p} \, \boldsymbol{e}_T + \Bar{H} \llbracket \boldsymbol{u} \rrbracket,\\
\end{aligned}
\end{equation}
where $\boldsymbol{e}_T$ is a unit vector in the direction of $\llbracket \boldsymbol{u} \rrbracket$.

\section{Implementation: architectures and training}
\label{sec:Architectures}

We propose two architectures to approximate the regularized strong discontinuity kinematics presented above. We consider the following features:
\begin{enumerate}
    \item An ANN that captures the regular behavior of the displacement field. This NN must be expressive enough to capture the behavior of the solid in the elastic domain. For the 1D case, we use a shallow ReLU-PINN to emulate an FEM approximation space (an explanation of this feature can be found in \cite{samaniego2020energy}). Meanwhile, a multilayer perceptron architecture with 4 hidden layers, each containing 10 neurons, is used for 2D examples, with the ReLU$(\mathbf{x})$ as the activation function. 
    \item An NN devised to capture the discontinuous part of the displacement field through a customized activation function $\Phi$.
\end{enumerate}

\noindent Thus, the NN approximation of the displacement field is constructed as
\begin{equation}
    \begin{split}
        \boldsymbol{u} \approx \boldsymbol{u}^* &= \boldsymbol{u}_{R} +  \boldsymbol{u}_{J}  , \\
        \Bar{\boldsymbol{u}} \approx \boldsymbol{u}_{R} &= NN(\mathbf{x}), \\
          \mathcal{H}_{\Gamma_{h}} \llbracket \boldsymbol{u} \rrbracket \approx \boldsymbol{u}_{J} &= \mathbf{w}\cdot\Phi(\mathbf{x}) , \\
    \end{split}
\label{eq:kin_decom}
\end{equation}
where $\boldsymbol{u}_{R}$ is the approximation of the continuous part of the displacement field and $\boldsymbol{u}_{J}$ is the approximation of the jump part. The field $\boldsymbol{u}_{J}$ includes the customized activation function $\Phi$. The vector $\mathbf{w}$ captures the jump on $\Gamma_h$ which, due to isochoricity of the plastic strain, develops in the direction tangential to $\Gamma$. The values of $\Phi$ are constrained between 0 and 1.

\begin{figure}[t]
\centering
\includegraphics[width=\textwidth]{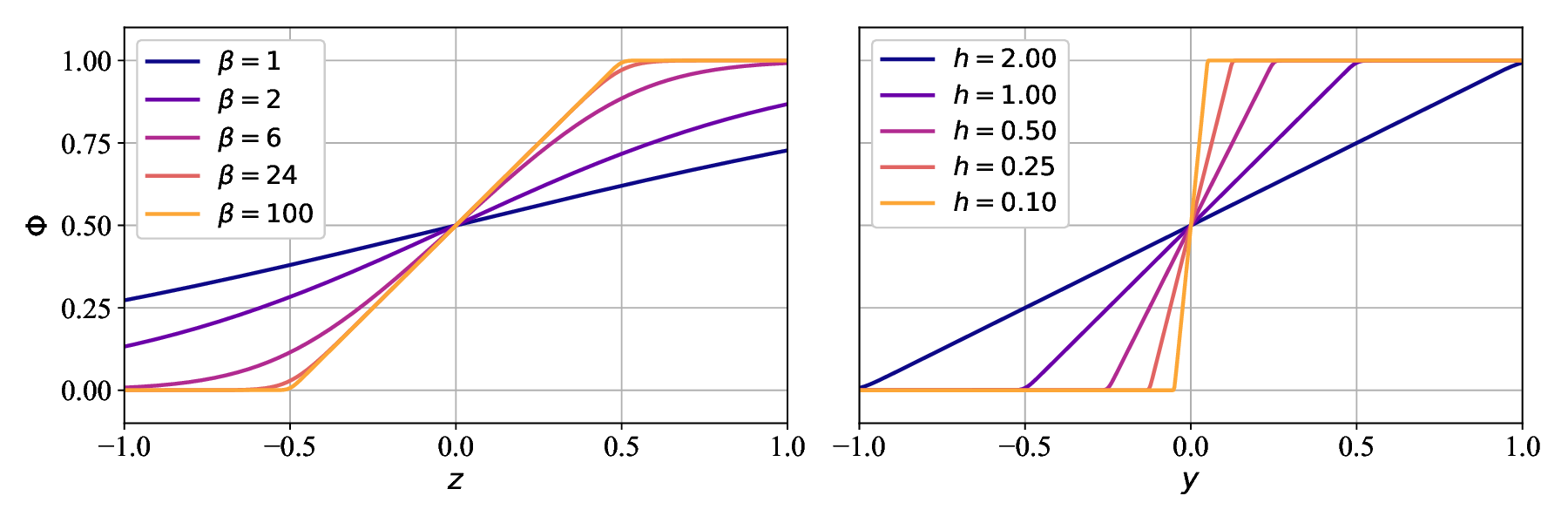}
\caption{Activation function. Left: effect of varying $\beta$ on $\Phi$. Right: effect of varying $h$ on $\Phi$, with $\beta$ fixed at 100.}
\centering
\label{fig_act}
\end{figure}

The activation function $\Phi$ (Figure \ref{fig_act}) is defined by
\begin{equation}
\begin{split}
    y &= \boldsymbol{x} \cdot \boldsymbol{n}, \\
    z &= \frac{y - y_{p}}{c},\\
    \Phi &= S(z + 0.5;\beta) - S(z - 0.5;\beta),
\end{split}  
\end{equation}
where $\boldsymbol{x}$ is the position vector, $\boldsymbol{n}$ is the unit vector normal to $\Gamma$ and is a trainable parameter, $y_{p}$ is also a trainable parameter that controls the position of the midpoint of the localization band, $c$ is a parameter that controls the width of the band, $\beta$ is a parameter that controls the sharpness of the band, and $S$ is the softplus function defined as $S(z,\beta)=\frac{1}{\beta}\ln{(1+e^{\beta z})}$.

\begin{algorithm}
\caption{Physics-Informed Neural Network (PINN) Training}\label{alg2d}
\begin{algorithmic}[1]
\State \textbf{Input:} Physics model (Variational formulation), boundary conditions
\State \textbf{Output:} Displacement field $\boldsymbol{u^{*}}$
\State Initialize neural network parameters $\boldsymbol{\theta}$ with the elastic state
\State Calculate the strain as $\mathbf{\epsilon^{*}}=\nabla^{s}\boldsymbol{u^{*}}$
\State Define the loss function $\mathcal{L}(\boldsymbol{\theta})$:
\begin{itemize}
    \item $\mathcal{L}_{\text{VF}}(\boldsymbol{\theta}) = \mathcal{W}[\boldsymbol{u^{*}}(\boldsymbol{\theta}),p(\boldsymbol{\theta})]$ \Comment{Variational Formulation loss}
    \item $\mathcal{L}_{\text{BC}}(\boldsymbol{\boldsymbol{\theta})} = \frac{1}{N_{bc}}\sum_{j=1}^{N_{bc}} \left\Vert \boldsymbol{u}^*(\boldsymbol{x}_j; \boldsymbol{\theta}) - \boldsymbol{u}_{bc}(\boldsymbol{x}_j) \right\Vert^2$ \Comment{Boundary condition loss}
    \item \textbf{if} $\mathcal{L}_{\text{BC}}(\boldsymbol{\theta}) > \lambda$ \textbf{then}
    \begin{itemize}
        \item[] $\mathcal{L}(\boldsymbol{\theta}) = \mathcal{L}_{\text{BC}}(\boldsymbol{\theta})$ \Comment{Loss considering only the boundary conditions}
    \end{itemize}
    \item \textbf{else}
    \begin{itemize}
        \item[] $\mathcal{L}(\boldsymbol{\theta}) = \mathcal{L}_{\text{VF}}(\boldsymbol{\theta}) +  \mathcal{L}_{\text{BC}}(\boldsymbol{\theta})$ \Comment{Total loss including energy of the solid}
    \end{itemize}
\end{itemize}
\State \textbf{while} not converged \textbf{do}
\begin{itemize}
    \item Compute the gradients of $\mathcal{L}(\boldsymbol{\theta})$ 
    \item Update $\boldsymbol{\theta}$ using AdamW optimizer
\end{itemize}
\State \textbf{end while}
\State Redefine the loss function $\mathcal{L}(\boldsymbol{\theta})$: \Comment{Done for 1D only}
    \begin{itemize}
        \item $\mathcal{L}(\boldsymbol{\theta}) = \mathcal{L}_{\text{VF}}(\boldsymbol{\theta}) +  \mathcal{L}_{\text{BC}}(\boldsymbol{\theta})$ \Comment{Loss considering both the energy and BC loss}
    \end{itemize}
\State \textbf{while} not converged \textbf{do} \Comment{Done for 1D only}
\begin{itemize}
    \item Compute the gradients of $\mathcal{L}(\boldsymbol{\theta})$ 
    \item Update $\boldsymbol{\theta}$ using L-BFGS optimizer
\end{itemize}
\State \textbf{end while}
\State Return the trained neural network model
\end{algorithmic}
\end{algorithm}

Regarding the training process, the following steps are followed (algorithm~\ref{alg2d}):
\begin{enumerate}
    \item The parameters of the neural network are initialized starting from the last elastic state, \textit{i.e.}, the last possible displacement without localization. These parameters are used in the subsequent loading states via transfer learning
    \item The energy of the solid is computed from Equation \eqref{energyfunct_plasticity} and also, for the loss function, the error of the boundary conditions is considered.
    \item If the error of the boundary conditions is greater than a certain threshold $\lambda$, the loss function is calculated only considering this error; otherwise, the loss function incorporates the energy of the solid.
    \item The loss function for the 1D example is minimized using the AdamW optimizer for 3000 epochs, followed by a refinement with the L-BFGS optimizer. For the 2D example, the loss function is minimized using the AdamW optimizer for 5000 epochs.
\end{enumerate}
 
In order to optimally explore the solution space of the NN, a learning rate schedule is used to adapt the way the AdamW optimizer varies at each step: at the beginning of the process, there are larger ``steps" to avoid possible local minima or saddle points, where the loss function might get stuck. In this paper we use the \texttt{CosineDecay()} learning schedule, implemented in TensorFlow.

\section{Numerical experiments}
\label{sec:results}

This section presents results for 1D and 2D test cases. The fundamental idea is to show that the proposed methodology is able to accurately reproduce the solution of the variational problem. Given that the examples provided can be solved analytically, we are able to show that the NN solution is, modulo some numerical errors, the solution of the original mathematical problem. This shows that the approximation space given by the proposed architectures contains the exact solution for the cases considered here, which suggests that appropriate approximation spaces can be devised within the framework of PINNs for solving strain localization problems with sharp discontinuities. 

The section is organized as follows. The results for the displacement field are shown for several loading stages, both for 1D and 2D examples. Then, for each case, the details of how the solution was obtained are analyzed and discussed.

\subsection{1D Example}
\label{sec:1d_example}

\begin{figure}[t]
\centering
\includegraphics[width=0.7\textwidth]{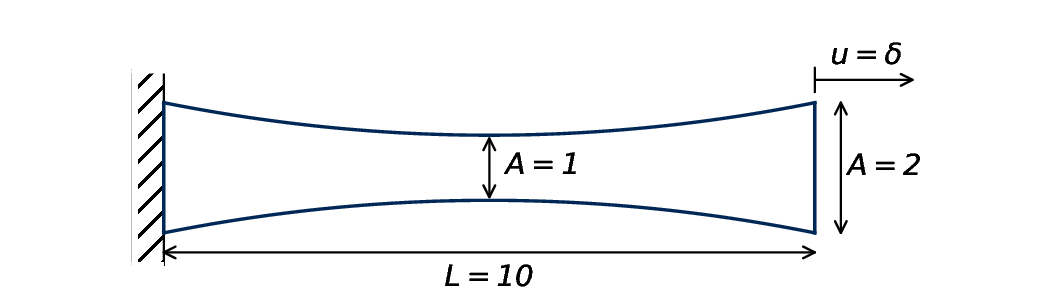}
\caption{Geometry configuration of the 1D problem.}
\centering
\label{fig_1Dgeom}
\end{figure}

The first example consists of a 1D bar whose domain is the open interval $ \Omega=(0,L)$, where $L=10$ is the length of the bar. The bar is fixed at one end ($x=0$), while, at the other end ($x=L$), the displacement is equal to $\delta$ (\textit{i.e.}, $u(L)=\delta$). The bar has a variable cross-sectional area in order for the problem of determining the location of the localization band to be uniquely defined. Specifically, the cross-sectional area $A(x)$ has a parabolic variation, with a minimum value of 1 at the midpoint and maximum values of 2 at the ends, as shown in Figure \ref{fig_1Dgeom}. Then, considering homogeneous material properties, the band is expected to localize at the middle of the bar. 

\begin{figure}[t]
\centering
\includegraphics[width=\textwidth]{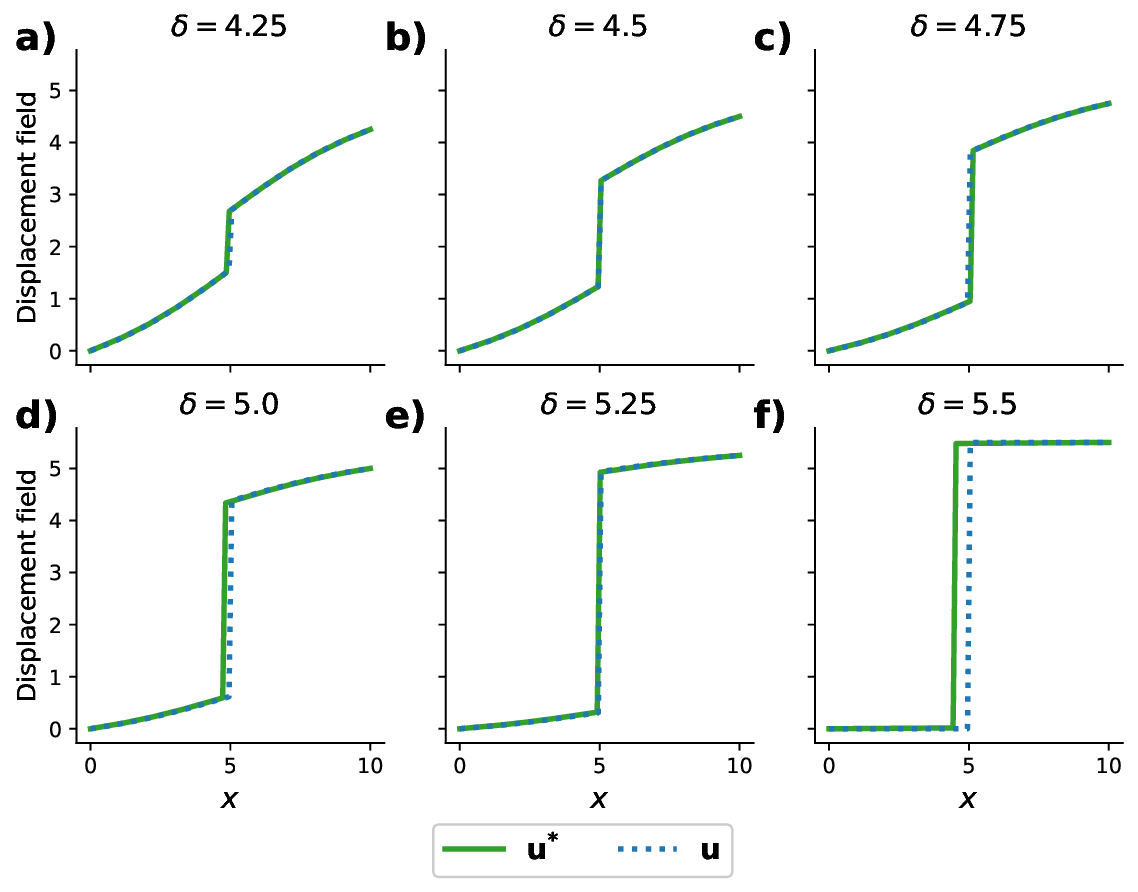}
\caption{Evolution of the displacement field for increasing prescribed displacement, showing the NN-predicted solution (solid green lines) and the expected analytical solution (dotted blue lines).}
\centering
\label{fig01}
\end{figure}

\begin{figure}[t]
\centering
\includegraphics[scale=0.75]{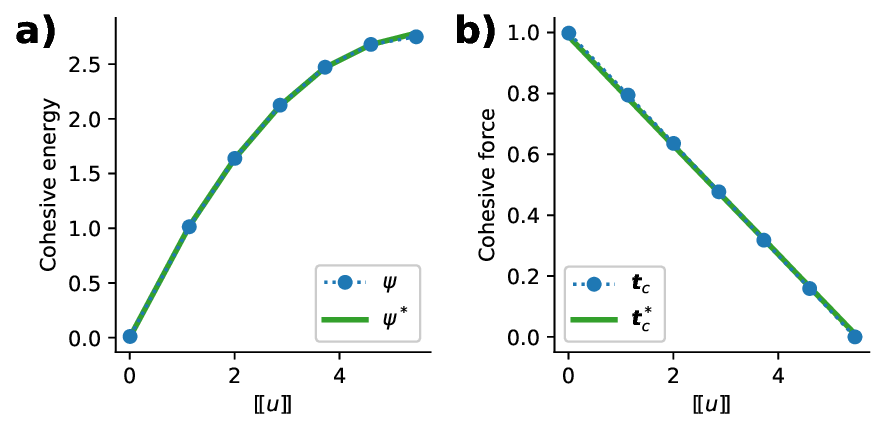}
\caption{Induced cohesive energy and cohesive force, showing the NN-predicted solution (solid green lines) and the expected analytical solution (dotted blue lines).}
\centering
\label{fig02}
\end{figure}

\begin{figure}[h!]
\centering
\includegraphics[width=\textwidth]{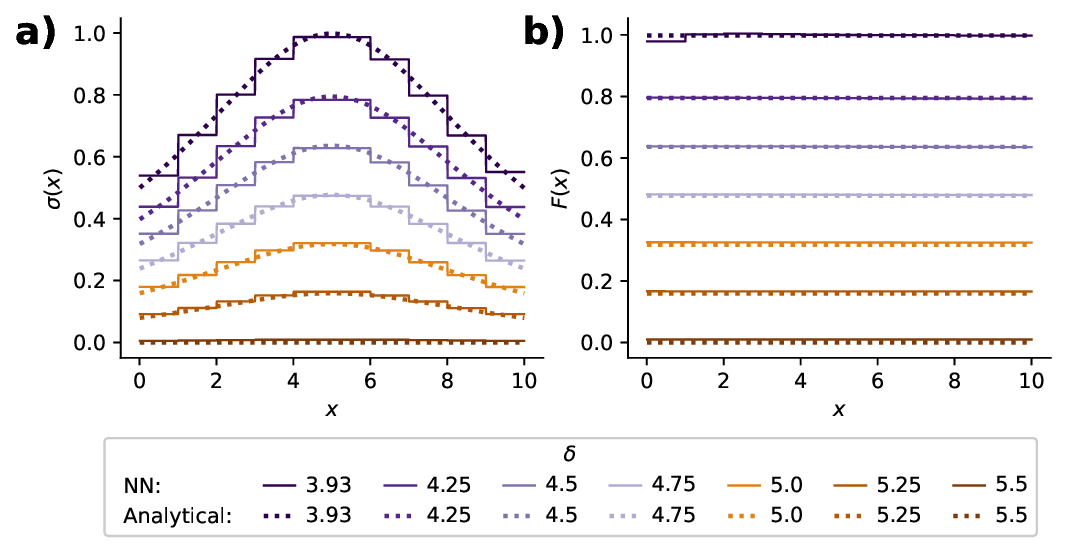}
\caption{Stress and force along the bar as the loading process evolves.}
\centering
\label{fig03}
\end{figure}

The material of the bar is characterized by the following parameters: $\sigma_p=1$, $E=2$ and $\Bar{H}=-2/11$. Notice that, in the regularized strong discontinuity approach used here, the bandwidth $h$ is not a material parameter given that, in principle, $h$ can take any sufficiently small value, and the plastic evolution is uniquely determined by the intrinsic softening modulus $\Bar{H}$.

In this example, we use 1001 collocation points uniformly distributed along the bar domain. For the regular part of the displacement field, the NN used emulates an FEM approximation space with 11 equally spaced FEM-like nodes. The parameters that describe the position of the band and the magnitude of the jump were initialized at 0.4 and 0.0, respectively.   

Figure \ref{fig01} shows that the predicted displacement field for various prescribed displacements $\delta$ closely matches the analytical solution. The regular part of the displacement has been approximated using 11 FEM-like nodes, which amount to a piecewise linear approximation. The NN also demonstrated a strong capability in approximating the jump in the displacement field. For all imposed displacements $\delta$, the training process led to accurate estimation of the jump magnitude within the localization band. In addition, the parameter describing the location of the band was effectively learned, with a noticeable difference with respect to the analytical solution occurring only for $\delta=5.5$. This shows the presence of the solution within the loss function landscape.

\begin{figure}[ht]
\centering
\includegraphics[width=\textwidth]{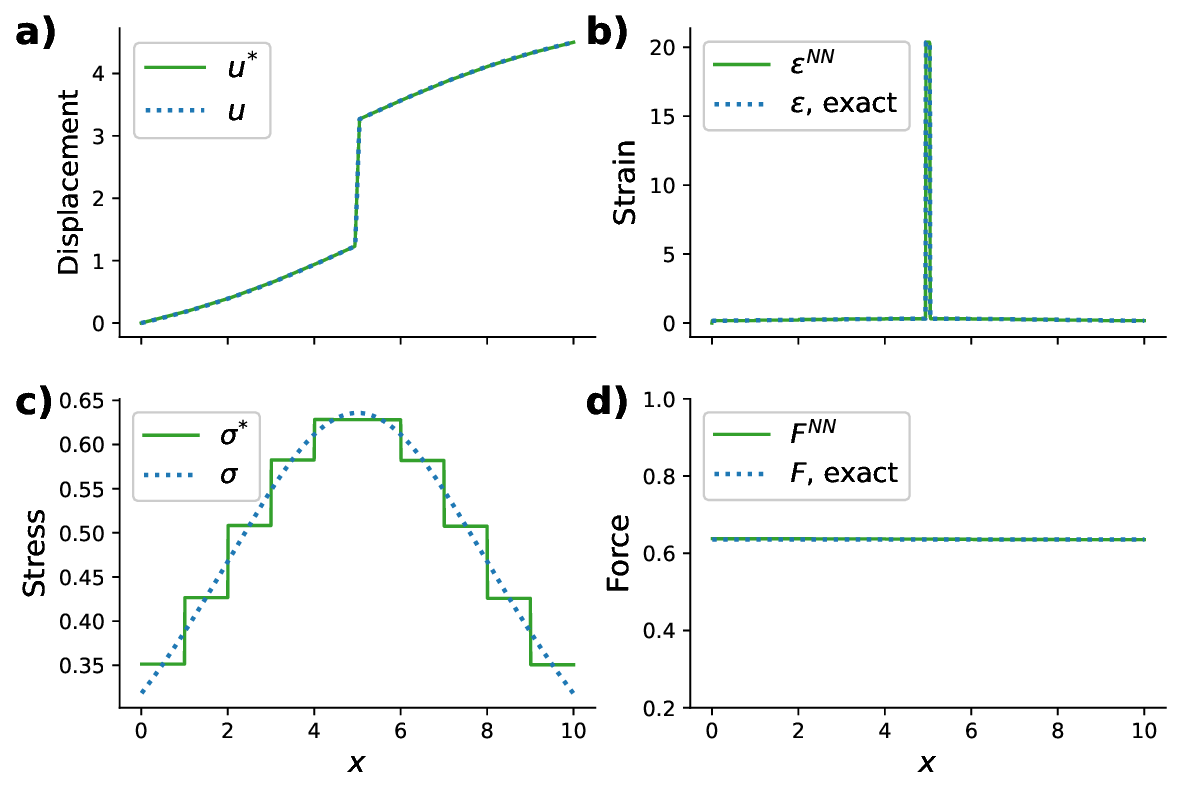}
\caption{Displacement, strain, stress and force along the bar as the loading process evolves, showing the NN-predicted solution (solid green lines) and the expected analytical solution (dotted blue lines).}
\centering
\label{fig04}
\end{figure}

\begin{figure}[ht]
\centering
\includegraphics[width=\textwidth]{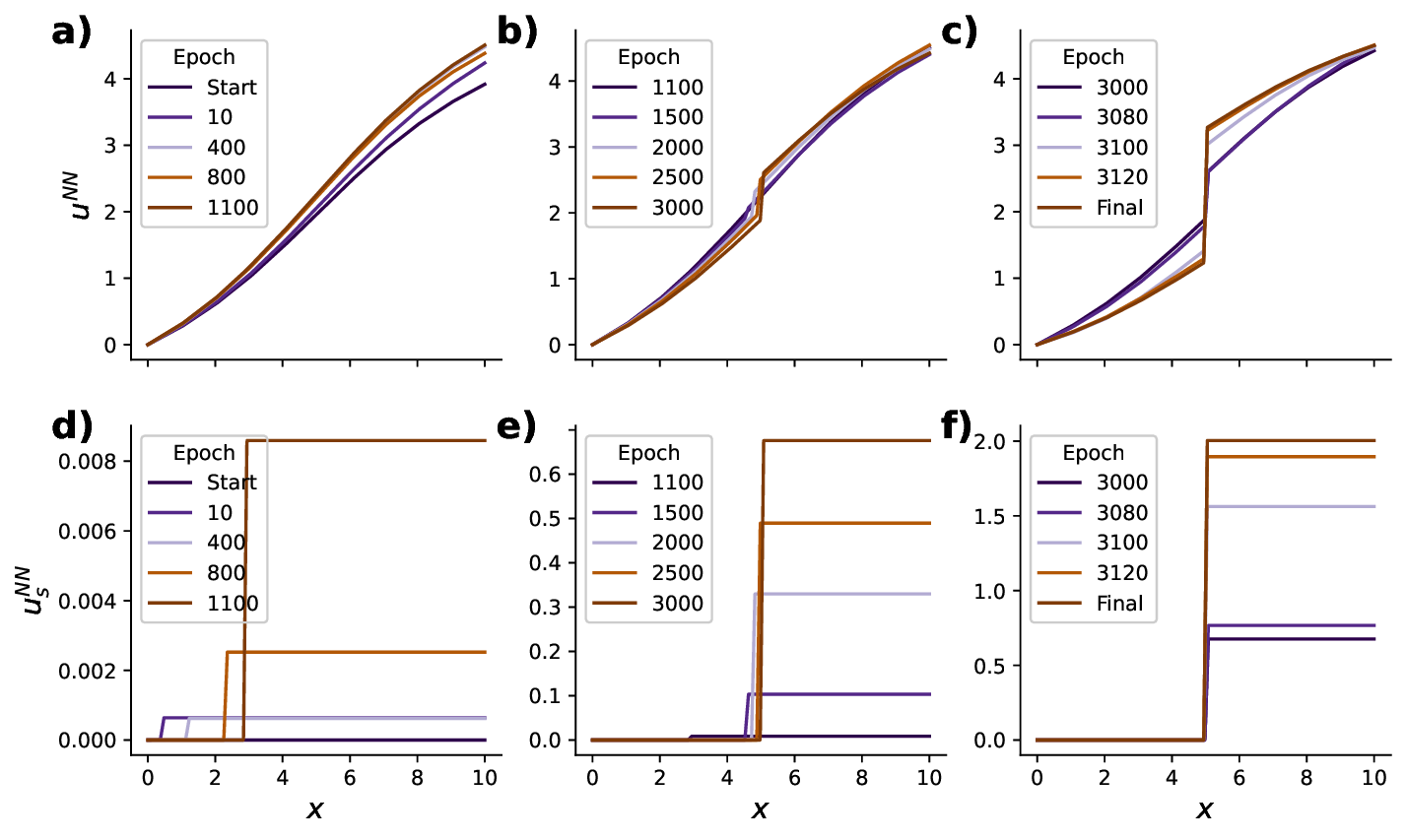}
\caption{Stress and force along the bar as the training process evolves.}
\centering
\label{fig05}
\end{figure}

Figure \ref{fig02} shows that the induced cohesive energy and force computed by the NN closely matches the analytical solution corresponding to the strong discontinuity approach. This shows that the regularized strong discontinuity kinematics and the variational statement of the BVP are able to capture the appearance of a cohesive band once softening triggers localization.

Figure \ref{fig03} illustrates the evolution of stress and force for increasing displacements. It is important to note that the stress field remains continuous even in the presence of strain localization within the band. Also, the force remains constant over the entire domain, as expected. For the case of $\delta=5.5$, where the loading process ends, both the stress and force approximate their actual value of 0, despite the minor error in the training of the band's position.

Here henceforth, we present an analysis using the specific case of $\delta=4.5$. Figure \ref{fig04} shows the displacement, strain, stress and force fields for this stage. This figure shows that the parameters describing the localization band were correctly learned. In Figure \ref{fig04} (b), both the location of the band and the magnitude of the jump are closely aligned with the analytical solution for the regularized strong discontinuity problem. 

The training process of the displacement field and the jump part of the displacement field are shown in Figure \ref{fig05} (a-c) and Figure \ref{fig05} (d-f), respectively. The latter is particularly important as it reveals how the localization band is trained using the AdamW and L-BFGS optimizers. The first 3000 epochs of training of the jump using the AdamW optimizer are plotted in Figure \ref{fig05} (d,e). There, we observe that the parameter describing the band's position is the first to be learned. In contrast, the magnitude of the jump continues to be trained. Then, during the training of the band using the L-BFGS optimizer, illustrated in Figure \ref{fig05} (f), the magnitude of the jump is finally learned. In fact, for all other cases of $\delta$ in the 1D example, a similar pattern was observed: the position of the localization band was primarily learned using the AdamW optimizer, while the magnitude of the jump continued to be trained using AdamW even when the band's position has almost reached its actual value, reaching its correct value with L-BFGS optimization.  

The computation time was approximately 2 minutes for each load state. All runs were performed on a personal computer with a 3.0 GHz AMD Ryzen 9 7845HX with Radeon Graphics (24 CPUs) processor and 16 GB of RAM.

\FloatBarrier

\subsection{2D Example}
\label{sec:2d_example}

The 2D case consists of a problem involving a square specimen of unit length subjected to simple shear. The specimen is subjected to Dirichlet boundary conditions, both at the bottom side, where it is fixed, and at the top side, where a tangential displacement of magnitude $\delta$ is applied.

The material parameters are $E=5.6$, $\nu=0.4$, $\Bar{H}=-1$, and $\sigma_p$, with the latter varying across the domain. Specifically, the yield stress $\sigma_p$ varies parabolically, with a minimum value of 0.75 along the horizontal line at the vertical center of the domain, and with a maximum value of 1 along both the top and bottom surfaces. This ensures that the localization band forms with a uniquely defined location and orientation. 

We use three different bandwidth values, $h\in\{0.05,0.1,0.2\}$, to show the objectivity of the solution with respect to the regularization parameter and, hence, with respect to the width of the localization band. In all cases, the domain contains a grid of 101 $\times$ 101 uniformly distributed collocation points along the $x$ and $y$ axis. 

The parameters describing the localization band are initialized as follows: the location of the band $y_p$ is set to 0.25; the orientation of the band is set to an angle of $\pi/2$ rads clockwise with respect to the horizontal axis; and the jump magnitude is set to zero for both the horizontal and vertical components of the displacement field.

\begin{figure}[b!]
\centering
\includegraphics[width=\textwidth]{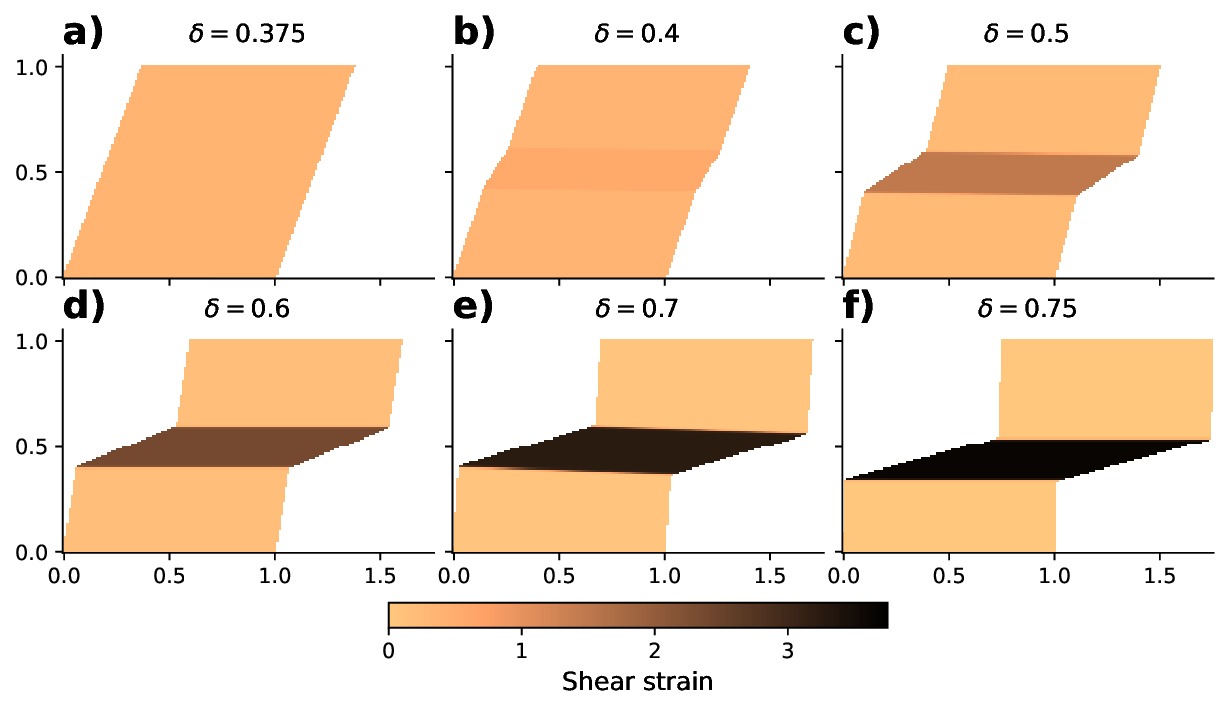}
\caption{Evolution of the displacement field with $h=0.2$ for increasing prescribed displacements.}
\centering
\label{fig06}
\end{figure}

\begin{figure}[h!]
\centering
\includegraphics[width=\textwidth]{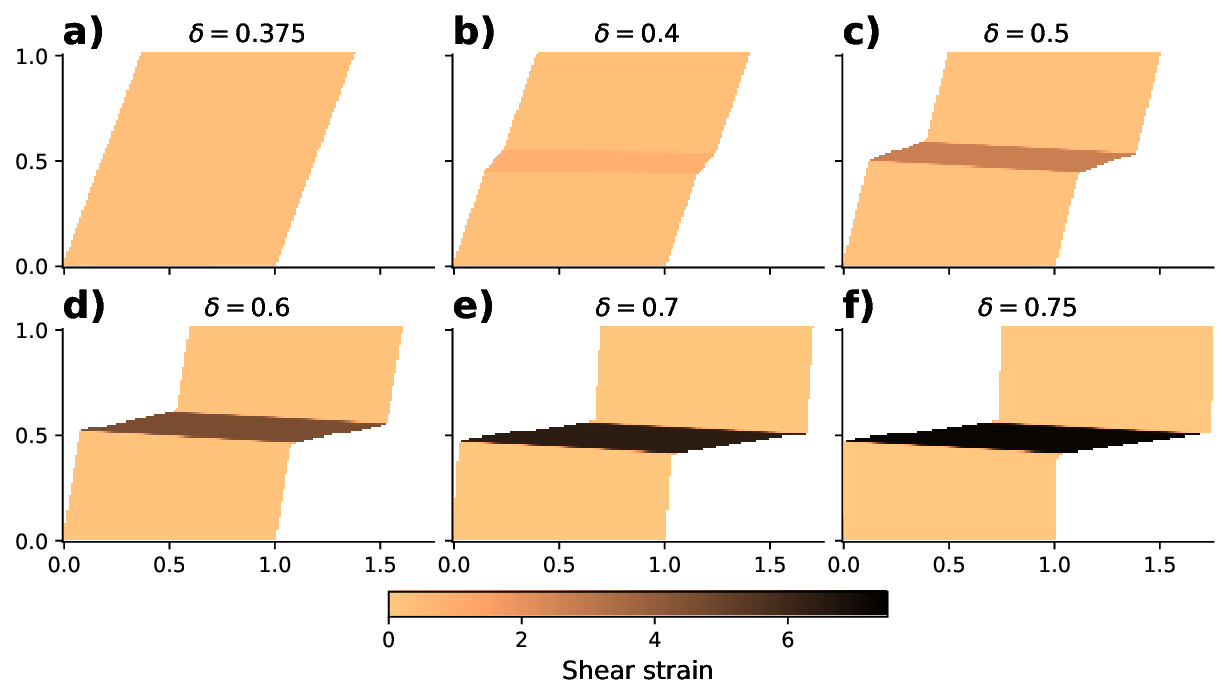}
\caption{Evolution of the displacement field with $h=0.1$ for increasing prescribed displacements.}
\centering
\label{fig07}
\end{figure}

\begin{figure}[h!]
\centering
\includegraphics[width=\textwidth]{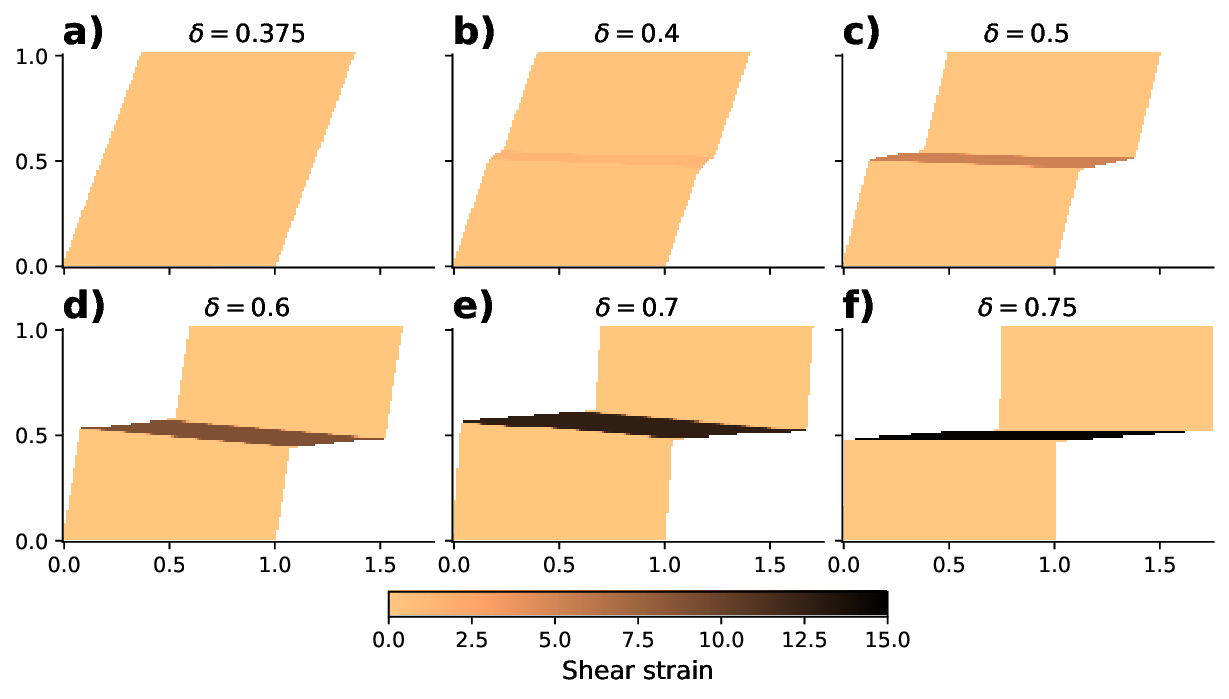}
\caption{Evolution of the displacement field with $h=0.05$ for increasing prescribed displacements.}
\centering
\label{fig08}
\end{figure}

\begin{figure}[t!]
\centering
\includegraphics[scale=0.75]{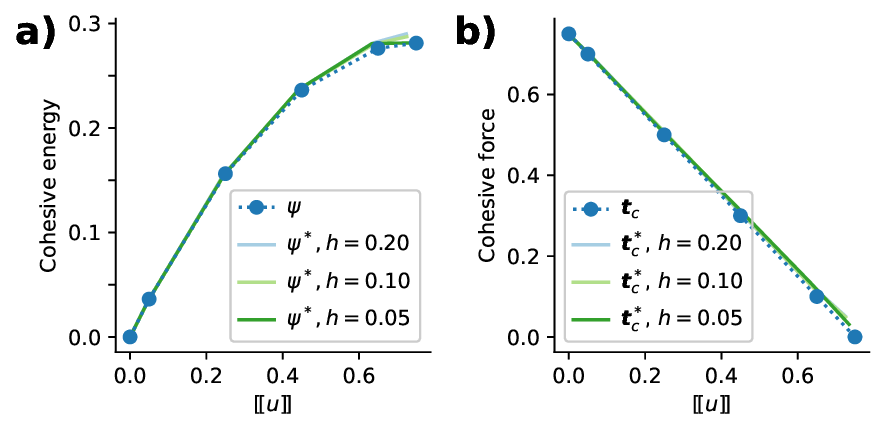}
\caption{Induced cohesive energy and cohesive force for different bandwidth values $h$ (solid lines), highlighting the objectivity of the solution and comparing the results with the expected analytical solution (dotted blue lines).}
\centering
\label{fig09}
\end{figure}

\begin{figure}[h!]
\centering
\includegraphics[scale=0.85]{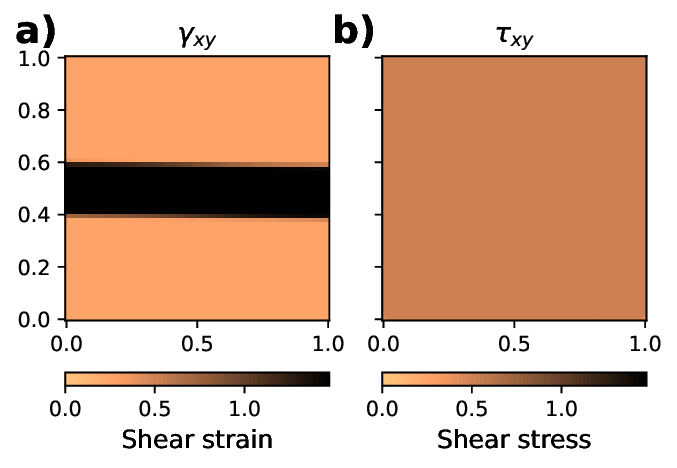}
\caption{Localized shear strain field and the corresponding uniform shear stress field.}
\centering
\label{fig10}
\end{figure}

\begin{figure}[t!]
\centering
\includegraphics[width=\textwidth]{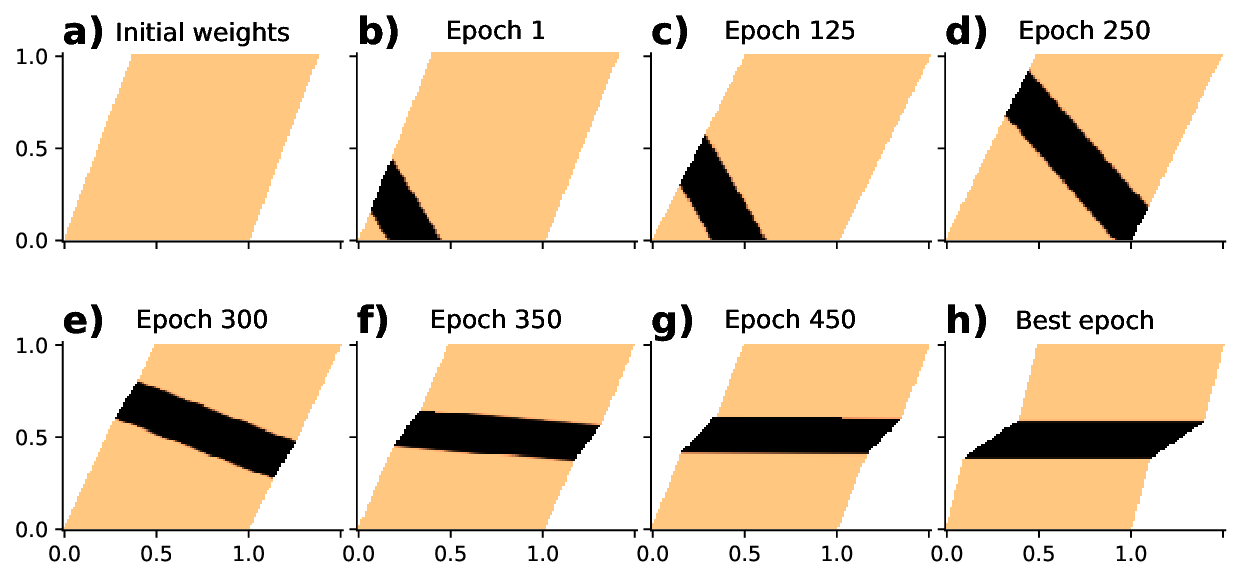}
\caption{Shear strain field as the training process evolves.}
\centering
\label{fig11}
\end{figure}

Figures \ref{fig06}, \ref{fig07} and \ref{fig08} show the deformed specimen for different prescribed displacements $\delta$, considering different bandwidths $h$ for each figure. We observe in all cases homogeneous strains at $\delta=0.375$ and localization triggered at $\delta=0.4$. As expected, the shear strains grow as the localization band shrinks with $h$, while the displacement magnitude remains invariant. Furthermore, notably, Figure \ref{fig09} illustrates that the present framework yields objective solutions, insensitive to the choice of bandwidth. In particular, we show that the cohesive energy and cohesive force are uniquely determined by the intrinsic softening parameter, as expected from Equations \eqref{cohesive_energy} and \eqref{cohesive_force}. Moreover, the NN-predicted values closely match the expected analytical solution.

Additionally, we show that the method is able to capture the uniformity of the stress, although the strain is concentrated in the band. This result can be observed in Figure \ref{fig10}.  Figure \ref{fig11} further illustrates the training process. It is especially remarkable that both the direction and the position of the band are captured by means of trainable parameters. 

The computation time was approximately 9 minutes for each load state. All runs were performed on a personal computer with a 3.0 GHz AMD Ryzen 9 7845HX with Radeon Graphics (24 CPUs) processor and 16 GB of RAM.

\subsection{Sensitivity analysis}

We consider the effect of varying several hyperparameters with respect to the numerical experiments performed in the previous sections. The main focus of this article is to explore the possibility of using energy minimization to deal with strain localization as a strong discontinuity. However, in practice, once the discretization of the energy functional using ANNs is performed, we are left with an optimization problem that can be certainly challenging.

\subsubsection{Number of collocation points and degrees of freedom in the 1D example}
\begin{figure}[t]
\centering
\includegraphics[width=\textwidth]{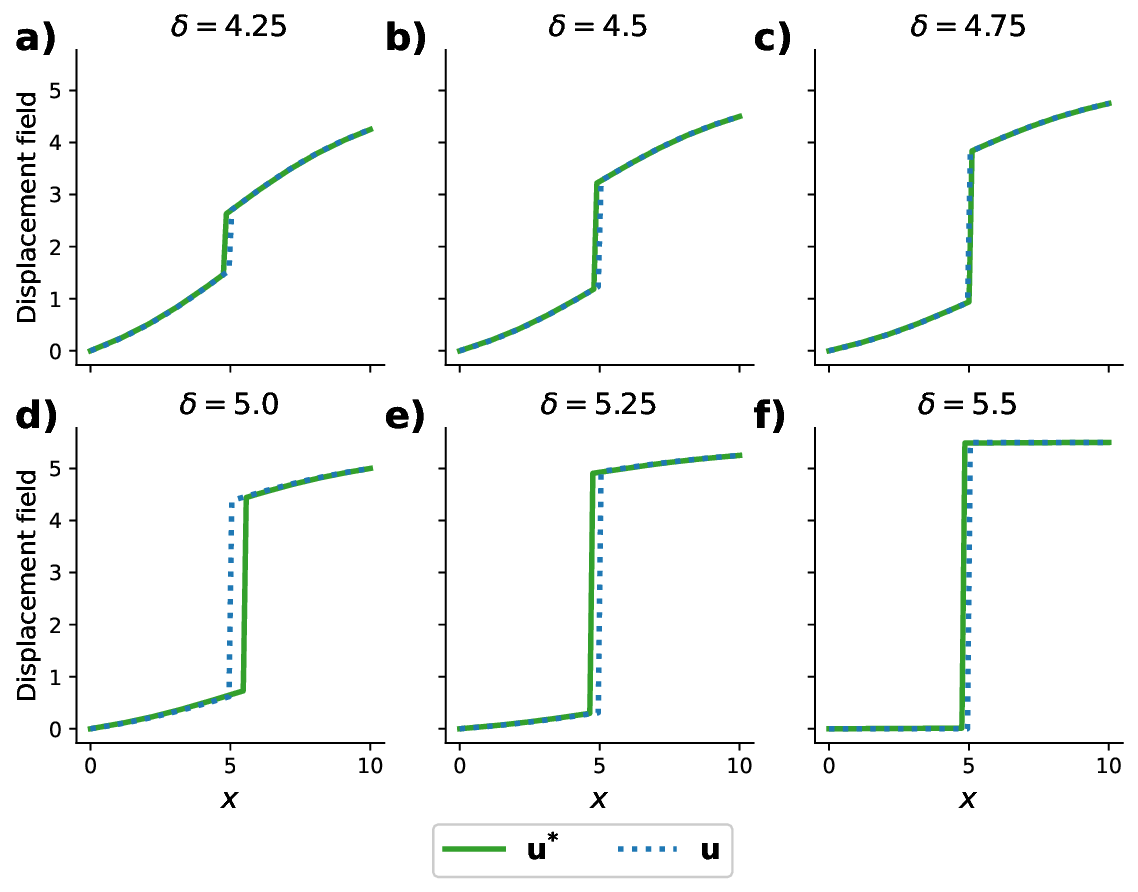}
\caption{Evolution of the displacement field for increasing prescribed displacement, showing the NN-predicted solution (solid green lines) and the expected analytical solution (dotted blue lines) when using 901 collocation points and 11 FEM-like nodes.}
\centering
\label{fig12}
\end{figure}

\begin{figure}[t]
\centering
\includegraphics[width=\textwidth]{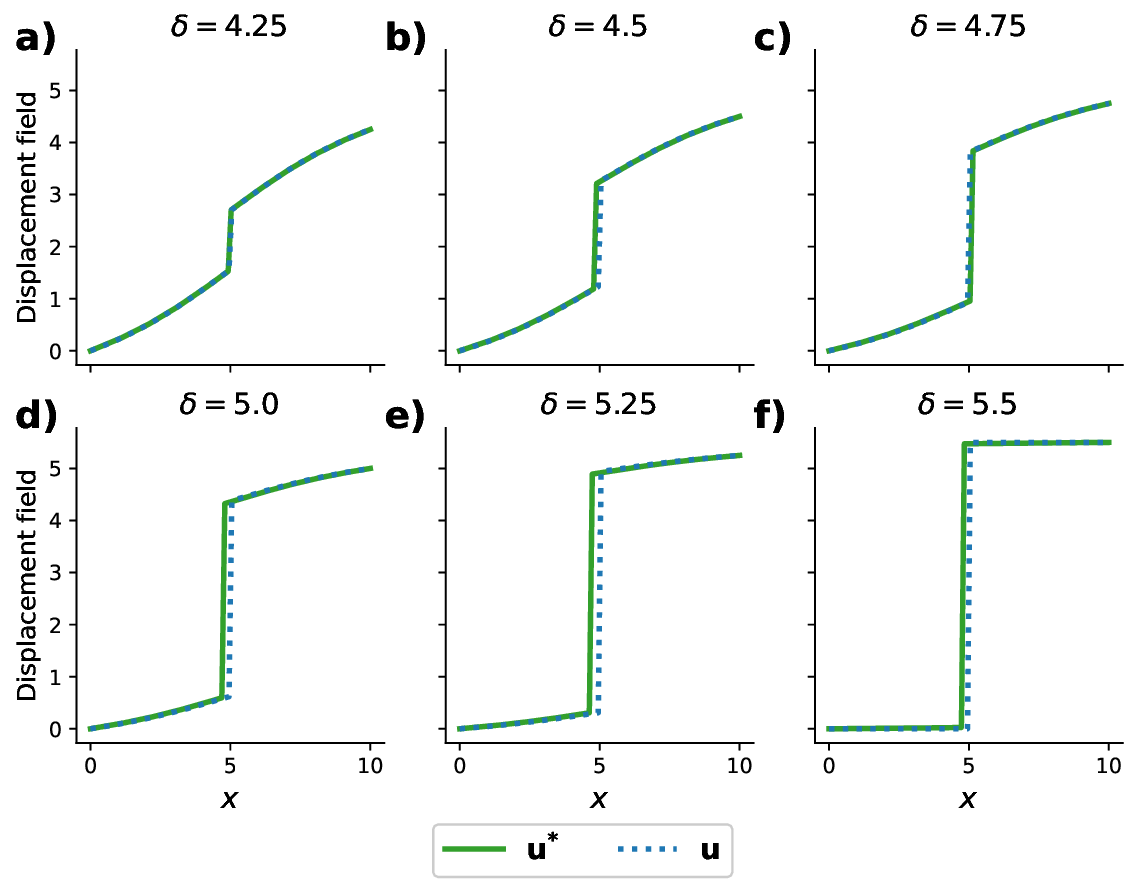}
\caption{Evolution of the displacement field for increasing prescribed displacement, showing the NN-predicted solution (solid green lines) and the expected analytical solution (dotted blue lines) when using 1101 collocation points and 11 FEM-like nodes.}
\centering
\label{fig13}
\end{figure}

A sensitivity analysis was performed concerning the number of collocation points and degrees of freedom for the 1D bar. First, we examined the effect of varying the number of collocation points. Specifically, the impact of using 901 and 1101 collocation points was evaluated compared to the 1001 points used in the previous section. The neural network architecture was unchanged, maintaining 11 degrees of freedom. Furthermore, all hyperparameters remain, including the learning schedule parameters and the threshold parameter $\lambda$ used for boundary conditions in the loss function. Additionally, the initialization of parameters defining the position of the localization band and the magnitude of the displacement jump were preserved. Figures \ref{fig12} and \ref{fig13} show the bar response for 901 and 1101 collocation points, respectively. The results demonstrate that the bar response is accurately captured in all cases, with the regular and jump displacement parts closely matching the analytical solution. The jump magnitude is estimated with high precision, while the position of the localization band is learned accurately, showing a maximum deviation of 0.51 for $\delta=5.0$ in the 901 collocation points analysis (Figure \ref{fig12}d).

\begin{figure}[t]
\centering
\includegraphics[width=\textwidth]{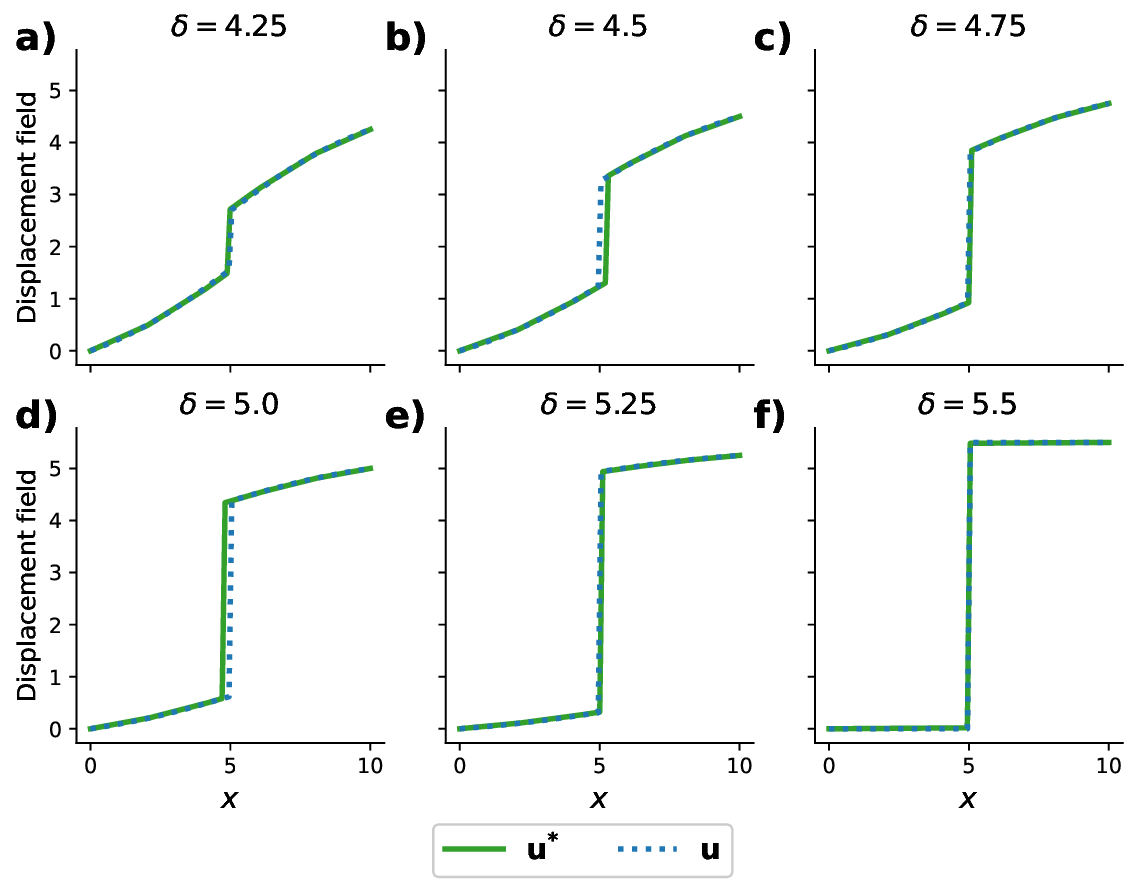}
\caption{Evolution of the displacement field for increasing prescribed displacement, showing the NN-predicted solution (solid green lines) and the expected analytical solution (dotted blue lines) when using 1001 collocation points and 6 FEM-like nodes.}
\centering
\label{fig14}
\end{figure}

\begin{figure}[t]
\centering
\includegraphics[width=\textwidth]{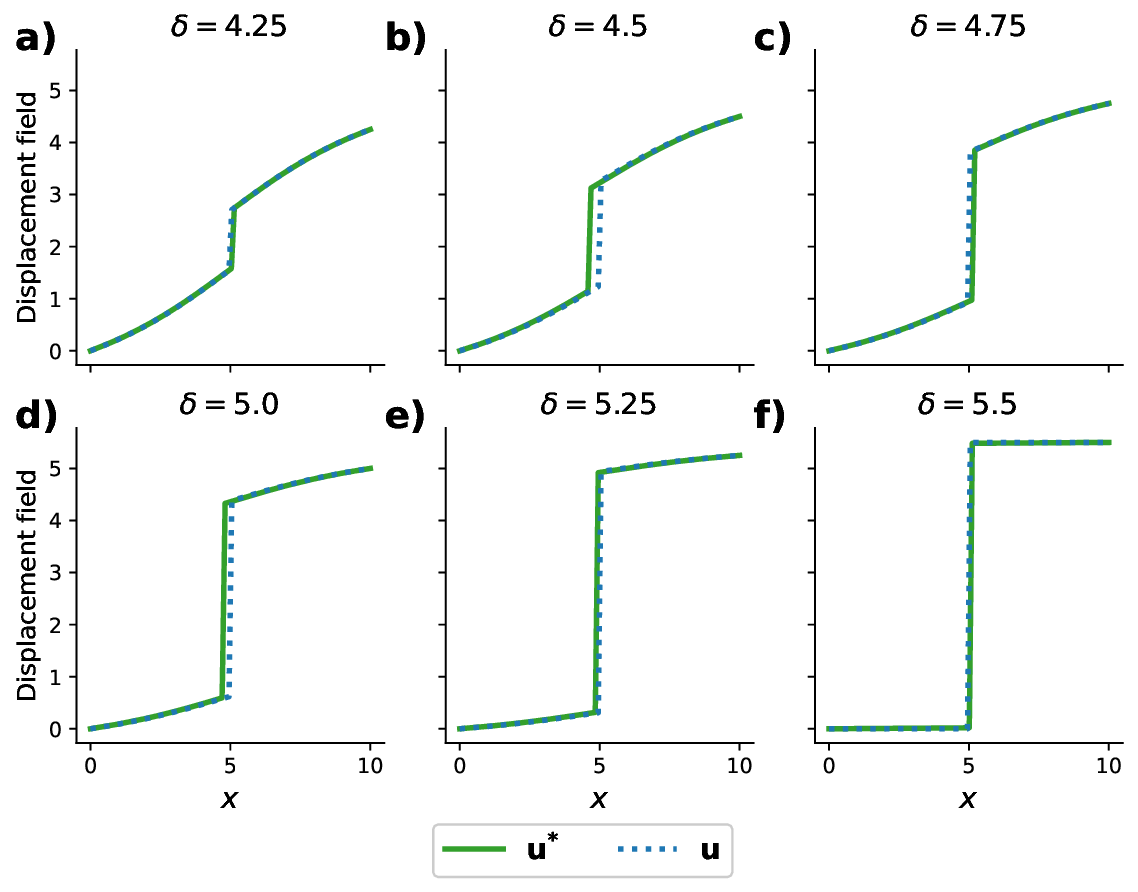}
\caption{Evolution of the displacement field for increasing prescribed displacement, showing the NN-predicted solution (solid green lines) and the expected analytical solution (dotted blue lines) when using 1001 collocation points and 21 FEM-like nodes.}
\centering
\label{fig15}
\end{figure}

Next, we examined the sensitivity to varying degrees of freedom in the neural network by changing the number of FEM-like nodes while keeping the number of collocation points fixed at 1001. In this case, we evaluated the response of the bars with 6 and 21 degrees of freedom, compared to the 11 degrees of freedom initially used in the previous section. Once again, the hyperparameter $\lambda$, the learning schedule, and the initialization of the parameters for the location of the localization band and the magnitude of the jump remain unchanged. Figures \ref{fig14} and \ref{fig15} represent the response of the bar for 6 and 21 FEM-like nodes, respectively. The results confirm that the location of the localization band and jump magnitude parameters are adequately learned in all cases. Overall, the bar response is well captured in most scenarios, with a maximum error of 0.37 observed in estimating the location of the localization band for $\delta=4.5$ in the sensitivity analysis with 21 FEM-like nodes (Figure \ref{fig15}b).

Regarding the training process, some key points must be highlighted. As stated earlier, the learning schedule parameters were kept unchanged, specifically those associated with the CosineDecay schedule used in the AdamW optimizer. Additionally, efforts were made to maintain consistency in the neural network training algorithm across all cases. However, in certain instances, adjustments to the training procedure were necessary to ensure accurate results. The following modifications were applied:

\begin{itemize}

\item Additional L-BFGS optimization round: In some cases, after completing the standard training schedule (3000 AdamW epochs followed by the L-BFGS optimization), an additional round of L-BFGS optimization was necessary. This additional step proved effective in accurately estimating the jump magnitude. The initial iterations of each L-BFGS optimization appeared to assist in escaping local minima or saddle points, thereby improving convergence and final results. Examples of this adjustment include $\delta=5.50$ in the sensitivity analysis with 901 collocation points (Figure \ref{fig12}f) and $\delta=4.75$ in the analysis with 1101 collocation points (Figure \ref{fig13}c).

\item Adjustment of AdamW epochs: The default 3000 epochs for AdamW were adjusted in specific cases before switching to the L-BFGS optimizer, particularly when the previous modification did not yield optimal results. For instance, using 2000 AdamW epochs produced better outcomes for $\delta$ values of 4.25, 4.50, and 5.25 with 6 FEM-like nodes (Figure \ref{fig14}a,b,e), as well as for $\delta=4.50$ with 1101 collocation points (Figure \ref{fig13}b). Similarly, improved convergence was achieved for $\delta=5.0$ with 6 FEM-like nodes using 1600 epochs (Figure \ref{fig14}d) and with 21 FEM-like nodes using 1900 epochs (Figure \ref{fig15}d). It is also important to note that, in all cases where this modification was applied, the energy-based loss after the adjusted number of AdamW epochs was higher than that after the default 3000 epochs. However, the subsequent L-BFGS optimization resulted in lower energy values. This indicates that carefully tuning the number of AdamW epochs can enhance the overall training process by promoting better convergence during the L-BFGS optimization.

\item Absolute value of the energy in the loss function: Although the training procedure was not strictly modified, an adjustment was made to calculate the energy component of the loss function. Specifically, the absolute value of the energy was used in some instances. Under the standard training process, the location of the localization band was generally well-approximated after 3000 AdamW epochs, with L-BFGS optimization refining the jump magnitude. However, for specific cases, the L-BFGS optimization caused the energy to diverge toward negative infinity, resulting in inaccurate predictions. This issue, documented in the literature \cite{rivera2022quadrature} for problems that are in principle even easier, was mitigated in our study using the energy's absolute value. This modification was used for $\delta$ values of 4.25, 4.75, and 5.25 in the sensitivity analysis with 21 FEM-like nodes (Figure \ref{fig15}a,c,e) and $\delta= 4.50$ in the sensitivity analysis with 901 collocation points (Figure \ref{fig12}b).

\end{itemize}

\begin{figure}[t]
\centering
\includegraphics[scale=0.85]{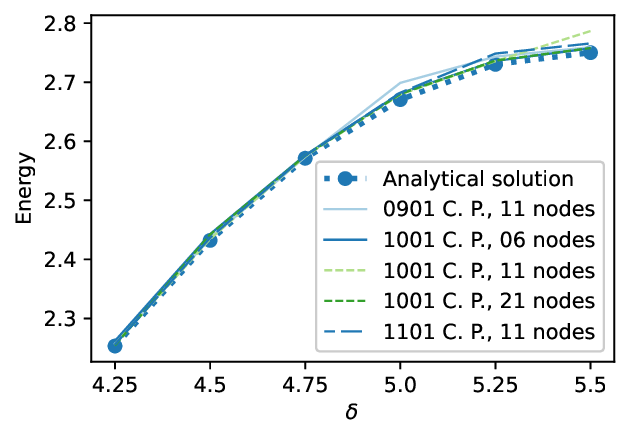}
\caption{Total energy of the bar for different $\delta$ values: analytical solution vs ANN prediction with varying numbers of degrees of freedom and collocation points.}
\centering
\label{fig_delta_vs_energy}
\end{figure}

It is worth mentioning that in the cases analyzed, only one of the mentioned modifications was applied at a time to keep deviations from the original training procedure to a minimum. Furthermore, these modifications resulted in solutions that not only closely matched the analytical displacement field, but also provided a satisfactory estimate of the total energy of the bar, as shown in Figure \ref{fig_delta_vs_energy}.  As shown in Figures \ref{fig12} to \ref{fig_delta_vs_energy}, the framework presented here produces satisfactory results regardless of the number of collocation points or the degrees of freedom. However, it highlights the importance of further investigation, particularly in the training process.

\subsubsection{Initial position and angle in the 2D example}

We have performed a study of the influence of certain hyperparameters for the 2D example. More specifically, we have studied the influence of the initialized position and direction of the localization band. For this purpose, values of 0.2, 0.5, and 0.8 were selected for the position, and values of 0, 15, 30, 45, 60, and 75 degrees were selected for the initial direction of the normal vector to the band. Then, we applied these 18 possible combinations to 3 quasi-static states, for an imposed displacement of 0.4, 0.55, and 0.7.

\begin{figure}[t]
\centering
\includegraphics[width=\textwidth]{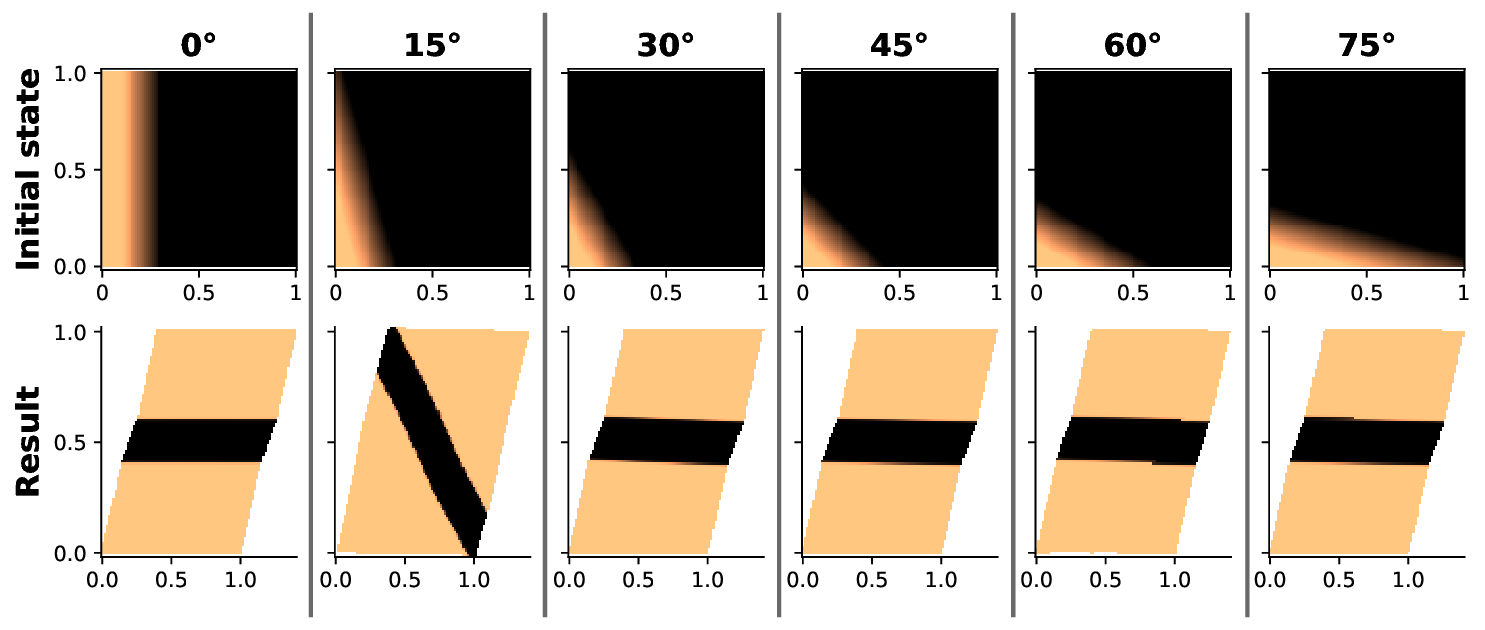}
\caption{Result of varying the initial orientation for an initial position of 0.2 and $\delta=0.4$.}
\centering
\label{fig16}
\end{figure}

\begin{figure}[t]
\centering
\includegraphics[width=\textwidth]{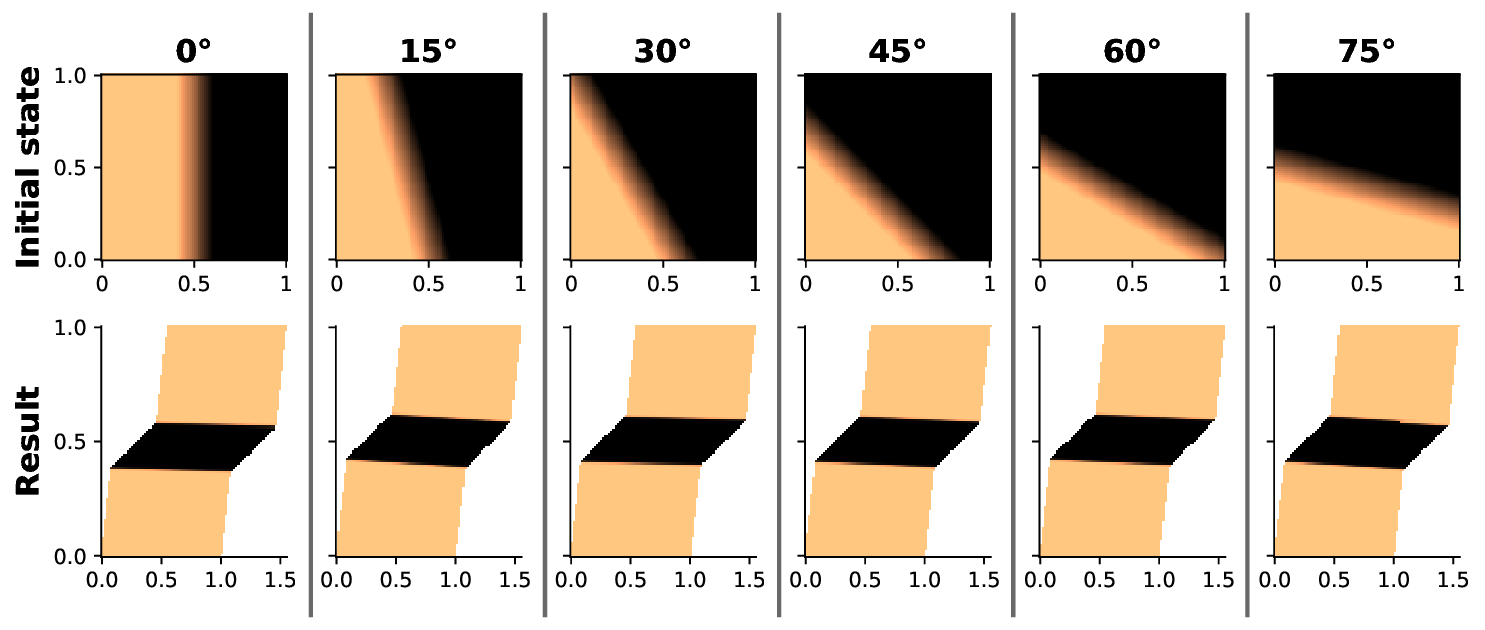}
\caption{Result of varying the initial orientation for an initial position of 0.5 and $\delta=0.55$.}
\centering
\label{fig17}
\end{figure}

\begin{figure}[t]
\centering
\includegraphics[width=\textwidth]{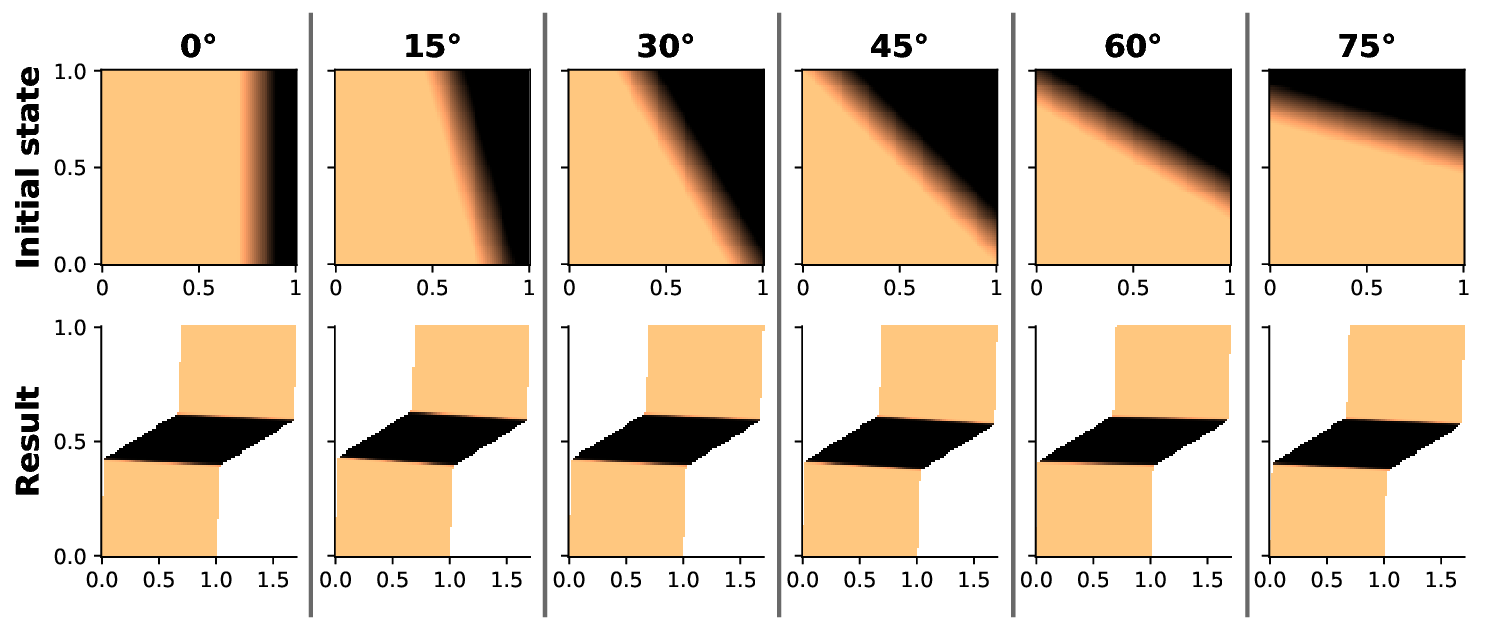}
\caption{Result of varying the initial orientation for an initial position of 0.8 and $\delta=0.7$.}
\centering
\label{fig18}
\end{figure}

The analysis of the results, illustrated in figures \ref{fig16}-\ref{fig18}, reveals that variations in both the initial position and orientation of the band have no significant impact on the final predictions of the network. In these figures, the upper row shows the initial location of the band; it is possible to observe how the angle of the normal to the band varies as well as its position. Only minor deviations are observed between the resulting configurations after energy minimization.

Although occasional deviations from the actual solution are observed (as in figure \ref{fig16}-b), it is unlikely that these discrepancies are due to the initial band parameters themselves. Instead, it is more plausible that they are due to limitations in the optimization process, since most of the initial locations studied here result in solutions that are very close to the real one.
In this context, two critical factors underlying the training process must be considered:

\begin{itemize}
    \item The landscape of the loss function: using ANN entails dealing with a complex loss landscape, potentially featuring local minima or saddle points that can hinder the optimization process for certain cases.
    \item The optimization algorithm: navigating the loss landscape uses some optimization algorithm, which, combined with the complex landscape of the loss function, inherently influences the training outcome. The optimizer's performance is determined by several hyperparameters, all of which play a role in the convergence behavior. Hyperparameter tuning and the choice of optimization algorithms, such as Adam or L-BFGS, impact the ability to attain a global minimum.
\end{itemize}

With these considerations in mind, it is worth noting that the position and orientation of the band do not, in general, influence the ability of the network to accurately approximate the true solution significantly. Although some discrepancies are observed, they are best attributed to problems inherent to the optimization process, which is a very interesting research direction, but outside the scope of the present study.

\subsubsection{Varying the number of collocation points in the 2D example}

\begin{figure}[t]
\centering
\includegraphics[width=\textwidth]{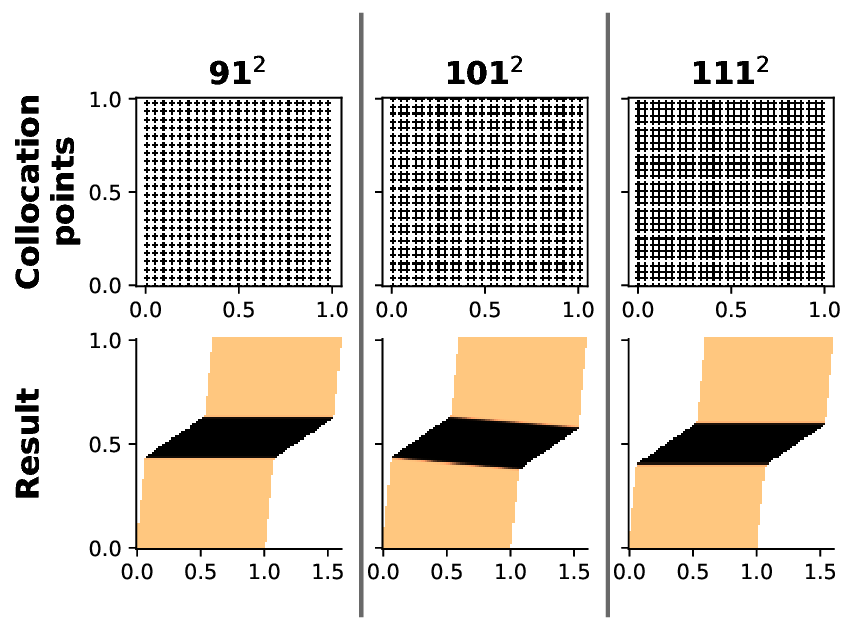}
\caption{Result of varying the number of collocation points for $\delta=0.6$.}
\centering
\label{fig19_collocations}
\end{figure}

To evaluate the influence of the number of collocation points on the simulation results, three cases were analyzed: 8281, 10201, and 12321 points. The corresponding results are presented in Figure \ref{fig19_collocations}. It can be observed that the impact of the number of collocation points on the results is not significant. Nevertheless, it is important to emphasize that, since the method relies on a globally evaluated energy functional, maintaining a sufficient number of collocation points is crucial to accurately compute the numerical integral.

\section{Conclusions}
\label{sec:conclusions}

We have used 1D and 2D numerical examples as proof of concept that plastic solids undergoing strain localization can be, in principle, modeled using energy minimization combined with regularized strong discontinuity kinematics to approximate the displacement field. In particular, we have been able to predict the onset and evolution of a localization band, including its position and the displacement jump magnitude, via trainable parameters in an ANN, addressing persistent challenges in finite element technologies endowed with sharp discontinuities. This result suggests that variational versions of PINNs hold promise for solving complex computational solid mechanics problems, such as materials with strain localization.

The present study sets the stage for subsequent developments on several fronts. Firstly, we emphasize that this study has not focused on developing a systematic and efficient numerical procedure but rather on proposing a potentially new paradigmatic approach to strain localization. Therefore, future work should focus on optimizing the computational cost, which is rather high for the training process, and streamlining the overall procedure. On the other hand, the idea of describing strong discontinuities via trainable ANNs can be applied to many cases beyond softening plasticity, \textit{e.g.}, frictional-dilatant plasticity or damage, and with more general loading conditions. Moreover, while we have shown feasibility for the onset and evolution of fixed localization bands, applying the proposed scheme to general boundary value problems requires handling crack/band propagation. A possible way forward is to develop multiscale techniques with ANN-based discontinuities propagating through a macroscopic mesh.

\section*{CRediT authorship contribution statement}
\label{sec:CRediT}
\textbf{Omar León}: Methodology, Software, Formal analysis, Investigation, Writing - Original Draft, Writing - Review \& Editing, Visualization. \textbf{Víctor Rivera}: Methodology, Software, Formal analysis, Investigation, Writing - Original Draft, Writing - Review \& Editing, Visualization. \textbf{Angel Vázquez-Patiño}: Methodology, Software, Writing - Review \& Editing, Visualization. \textbf{Jacinto Ulloa}: Validation, Formal analysis, Writing - Review \& Editing. \textbf{Esteban Samaniego}: Conceptualization, Methodology, Validation, Formal analysis, Writing - Original Draft, Writing - Review \& Editing, Supervision, Project administration.

\section*{Data availability}
No data was used for the research described in the article.

\section*{Acknowledgements}
\label{sec:acknowledgements}

A. V.-P. acknowledges to the VIUC for supporting him through “Conjunto de horas 1”.



 \bibliographystyle{elsarticle-num} 
 \bibliography{cas-refs}





\end{document}